\documentclass[10pt,twocolumn,letterpaper]{article}

\usepackage[algorithms]{wacv}      

\usepackage{graphicx}
\usepackage{amsmath}
\usepackage{amssymb}
\usepackage{booktabs}

\usepackage[accsupp]{axessibility}  

\usepackage{float}
\usepackage{caption}
\usepackage{subcaption}
\usepackage{booktabs}

\usepackage[pagebackref,breaklinks,colorlinks]{hyperref}

\usepackage[capitalize]{cleveref}
\crefname{section}{Sec.}{Secs.}
\Crefname{section}{Section}{Sections}
\Crefname{table}{Table}{Tables}
\crefname{table}{Tab.}{Tabs.}

\def\wacvPaperID{123} 
\def\confName{WACV}
\def\confYear{2024}

\begin{document}

\title{Amodal Intra-class Instance Segmentation: Synthetic Datasets and Benchmark}

\author{Jiayang Ao\\
The University of Melbourne\\
Parkville VIC 3010\\
{\tt\small jiayanga@student.unimelb.edu.au}
\and
Qiuhong Ke\\
Monash University\\
Clayton VIC 3800\\
{\tt\small Qiuhong.Ke@monash.edu}
\and
Krista A. Ehinger\\
The University of Melbourne\\
Parkville VIC 3010\\
{\tt\small kehinger@unimelb.edu.au}
}

\twocolumn[{%
\maketitle
\vspace*{-0.9cm}
\begin{figure}[H]
    \hsize=\textwidth 
    \centering
    \includegraphics[width=\textwidth]{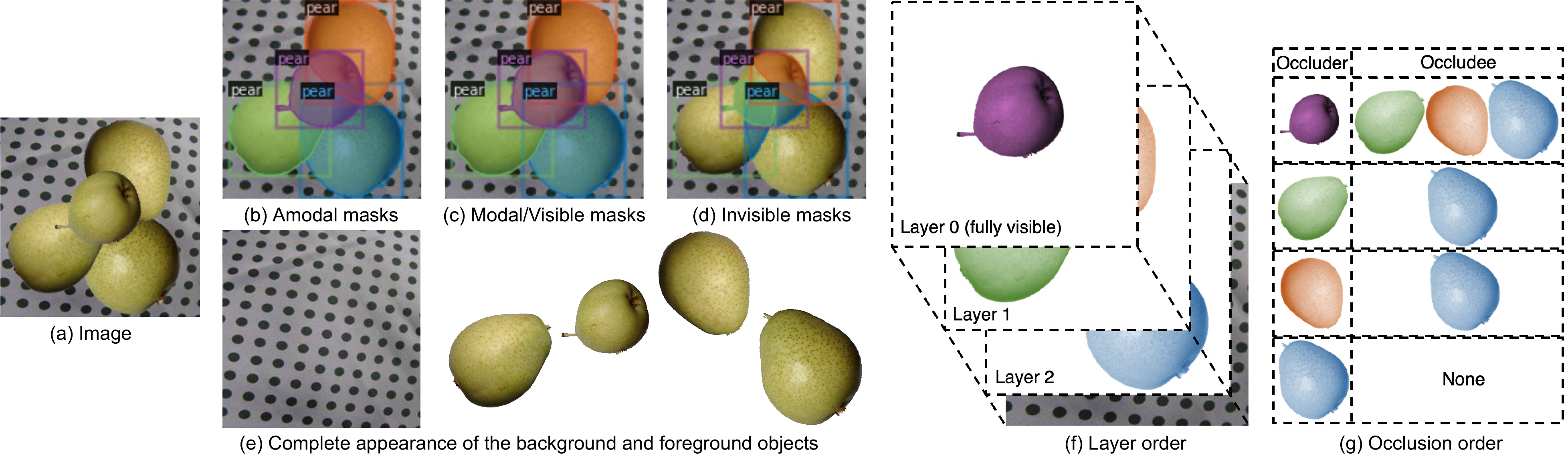}
    \caption{Overview of the proposed synthetic datasets. We build two challenging datasets of dense intra-class occlusion scenarios and provide detailed ground truth annotations of multiple types. 
    }
    \label{dataset overview}
\end{figure}
}]

\begin{abstract}
Images of realistic scenes often contain intra-class objects that are heavily occluded from each other, making the amodal perception task that requires parsing the occluded parts of the objects challenging. Although important for downstream tasks such as robotic grasping systems, the lack of large-scale amodal datasets with detailed annotations makes it difficult to model intra-class occlusions explicitly. This paper introduces two new amodal datasets for image amodal completion tasks, which contain a total of over 267K images of intra-class occlusion scenarios, annotated with multiple masks, amodal bounding boxes, dual order relations and full appearance for instances and background. We also present a point-supervised scheme with layer priors for amodal instance segmentation specifically designed for intra-class occlusion scenarios\footnote{The proposed datasets can be downloaded from this link: https://github.com/saraao/amodal-dataset.}. Experiments show that our weakly supervised approach outperforms the SOTA fully supervised methods, while our layer priors design exhibits remarkable performance improvements in the case of intra-class occlusion in both synthetic and real images.
\end{abstract}

\section{Introduction}

Perceiving an entire object, even if it contains invisible segments, is a task that can be easily handled by the human vision system \cite{kanizsa1979organization, chen2016amodal}, but it is still a challenge for computer vision. This capability, known as amodal perception, is crucial for vision tasks that require reasoning about occluded scenes to ensure reliability and safety, such as robot grasping systems \cite{back2022unseen} and autonomous driving \cite{qi2019amodal}. With the aim to mimic human amodal perception, amodal instance segmentation extends the traditional instance segmentation by additionally segmenting the occluded regions of each instance. While considerable progress has been made in perceiving visible regions for computer vision \cite{kirillov2020pointrend, cheng2022pointly}, amodal instance segmentation has received limited attention.

We tackle a problem not addressed by the previous works: namely, amodal instance segmentation with heavy intra-class occlusion. How to handle the images with intra-class instance occlusion is one big challenge in amodal instance segmentation. Due to the similarity between objects in the same class, intra-class instance segmentation becomes more challenging than inter-class instance segmentation, as class-specific features appear almost everywhere near the occlusion boundary \cite{follmann2019learning}. In particular, perceiving the complete range of entities, including the hidden parts, can be extremely difficult for the machine when multiple objects of the same class partially overlap within the same region of interest (ROI). In practice, however, there are many scenes with dense intra-class occlusion, such as warehouses and agricultural sites where robots work. In spite of its importance for many downstream tasks, the lack of relevant amodal datasets with detailed annotations hinders explicit learning and modeling of intra-class occlusions.

Our new datasets are valuable to the novel and under-studied problem of intra-class occlusion, unseen in previous amodal datasets. Constructing large-scale real datasets with amodal annotations for dense intra-class object occlusions would be costly and subject to bias and inaccuracies arising from manual annotation, thus synthetic datasets that produce accurate ground truth data are more suitable for this problem. 
Our two synthetic amodal datasets contain a total of over 267K images of intra-class occlusion scenarios with a wide variety of annotation types generated by the automatic data synthesis process. For each instance, the annotation provides modal/visible mask (Fig.~\ref{dataset overview}c), invisible mask (Fig.~\ref{dataset overview}d), and an amodal mask that combines visible and invisible masks (Fig.~\ref{dataset overview}b), integral appearance image (Fig.~\ref{dataset overview}e), layer and occlusion orders (Fig.~\ref{dataset overview}f and Fig.~\ref{dataset overview}g). The rich annotation types mean that 
our proposed datasets not only work for amodal instance segmentation, but also contribute to other closely related amodal completion tasks, such as order perception of occluded objects and amodal appearance completion that aims to infer the texture of invisible parts of objects. 

Using the new datasets to simulate point-based ground truth, we propose a point-supervised method with layer priors for the amodal instance segmentation task, which excels in handling intra-class occlusion. Our large-scale synthetic datasets enable fair comparisons and more thorough training of different methods on accurate ground truth data, and contribute to models’ performance on real images. Experiments show our weakly supervised method outperforms the other state-of-the-art (SOTA) weakly supervised \cite{cheng2023boxteacher, cheng2022pointly, tian2021boxinst} or even fully supervised methods \cite{cai2018cascade, follmann2019learning, he2017mask, kirillov2020pointrend, lee2020centermask, Tran_2022_BMVC} on intra-class occlusion in both synthetic and real images.

Our point-supervised method is well-suited for scenarios with limited or expensive full annotations, making it generalizable to real images. This is a crucial motivation for our approach, as full annotations can often be a limiting factor in many real-world applications. Thus, our method provides an effective benchmark for the challenging intra-class occlusion problem, as well as a practical and scalable solution that can be applied in real-world settings.

The main contributions of this paper are as follows:

\begin{itemize}
\item To the best of our knowledge, this is the first work that focuses on amodal instance segmentation of intra-class overlaps. To advance this novel task, we build the largest dataset of amodal images to date.

\item This is the first work to utilize point-based supervision in amodal tasks and introduces a novel method to represent multilayer image structure, Layer Priors. The point-based annotation scheme simplifies ground truth collection for future amodal datasets, while importing layer priors shows great effectiveness and simultaneously empowers layer order perception. 

\item To evaluate the generalisability of our method, we further collect a real dataset to test the proposed model trained on the synthetic dataset. Experimental results on both the synthetic and real datasets demonstrate that our proposed weakly supervised approach outperforms the existing SOTA weakly supervised and fully supervised methods.
\end{itemize}

\section{Related Work}

\subsection{Amodal Instance Segmentation}
Amodal instance segmentation assigns pixel-level categorical labels to visible and invisible regions of occluded objects to achieve amodal shape perception. The methods of amodal instance segmentation can be divided into two types depending on the supervision manner. Most of the previous related work applied full supervision. Typical strategies involve the use of additional amodal branches \cite{qi2019amodal, follmann2019learning, hu2019sail}, or prior cues \cite{xiao2021amodal, zhou2021human} on networks designed for modal perception to achieve amodal instance segmentation. However, previous approaches that treat the input image as a single layer are prone to fail in ROIs containing intra-class instance occlusion \cite{ke2021deep}. Our approach of treating an image as multilayer rather than single layer \cite{cheng2022pointly} or bilayer \cite{ke2021deep} thus helps to account for more complex occlusions and also makes it possible to perform multilayer order perception.

Another type of method makes an effort to alleviate the need for supervision. Considering weak supervision is particularly valuable for amodal perception tasks, as per-pixel manual annotators are expensive and usually fail to provide trustworthy ground truths due to subjective assumptions about invisible regions \cite{ehsani2018segan}, while computer-generated synthetic data suffer from a domain gap with the real-world scene. Previous amodal approaches suggested mitigating supervision by artificially covering the occluder's mask \cite{zhan2020self} or boundary \cite{nguyen2021weakly} to generate pseudo-ground truth. Compared to pseudo-ground truth, the point-based weak supervision adopted in this paper provides more trustworthy ground truth for a data-driven approach. The point annotation scheme requires annotating only ten random points rather than the entire mask per object \cite{cheng2022pointly}, which is particularly beneficial for amodal instance segmentation as fewer annotations reduce the bias introduced by manual annotation and expensive pixel-level labeling costs.

\begin{table*}[t]
    \footnotesize
    \centering
    \begin{tabular}{l | c c c  c c c c c c c c c c }
        \toprule
        Dataset & \# Img  &  \# Instances & \ \begin{tabular}[c]{@{}l@{}}Occluded \\ \# Instances  \end{tabular}      & \begin{tabular}[c]{@{}l@{}}Avg. Occ. \\Rate \%\end{tabular} & Mask & Appearance & \begin{tabular}[c]{@{}l@{}}Layer\\Order  \end{tabular} & \begin{tabular}[c]{@{}l@{}}Occ.\\Order  \end{tabular} & \begin{tabular}[c]{@{}l@{}}Target\\group \end{tabular}  \\
        \midrule
        COCOA \cite{zhu2017semantic} & 5,073 & 46,314 &  28,106 & 18.8 & $\checkmark$  & $-$  &   $-$   &   $\checkmark$    &COCO  \\
        D2SA \cite{follmann2019learning} & 5,600 & 28,720 & 16,337 & 15.0 & $\checkmark$ & $-$  &   $-$ & $-$ & Groceries
        \\
        KINS \cite{qi2019amodal}      & 14,991              & 190,626     & 99,964    & 19.8       & $\checkmark$  & $-$  & $\checkmark$     &    $-$    &Vehicles         \\
        \midrule
        Intra-AFruit (Ours)  & \textbf{255,000}  & \textbf{819,856} & \textbf{521,772} & \textbf{30.6} &$\checkmark$ &$\checkmark$&$\checkmark$&$\checkmark$ & Fruits\\
        \bottomrule
    \end{tabular}
    \vspace*{-0.1cm}
    \caption{Summary of some related amodal datasets. The proposed Intra-AFruit provides the largest amount of data and the richest available annotation types and focuses on a new target group, fruits. COCO refers to object categories in the COCO dataset \cite{lin2014microsoft}.}
    \label{datasets}
\end{table*}

\subsection{Amodal Perception Datasets}
A major challenge associated with the study of amodal perception is the lack of large-scale amodal datasets. Popular large-scale datasets for instance segmentation \cite{lin2014microsoft, Cordts2016Cityscapes} focus only on visible regions but not the hidden parts. To serve the amodal task, a few amodal datasets have been created with additional annotations to include the invisible parts. COCOA \cite{zhu2017semantic} provides class-agnostic amodal segmentation of each annotated segment for thousands of images from COCO \cite{lin2014microsoft} using inferences from human annotators. COCOA-cls \cite{follmann2019learning} further annotates parts of images from COCOA \cite{zhu2017semantic} with class-specific amodal instance masks. As a small dataset with a large number of object categories tends to lead to model overfitting, some amodal datasets focus only on vehicles or humans \cite{yan2019visualizing, zhou2021human, mohan2022amodal}. Although certain datasets feature scenes with vehicle occlusion \cite{qi2019amodal, breitenstein2022amodal}, the vehicles in these scenes often span multiple classes (e.g., car, van, truck, tram) and may be intermingled with human subjects, hence failing to serve the specific purpose of investigating intra-class occlusion issues. There is an urgent need for amodal datasets that explicitly support the study of intra-class occlusion.

Our dataset Intra-AFruit is explicitly built for the amodal perception of intra-class occlusion of fruits and vegetables as it facilitates wide-ranging applications, such as automated harvesting, robotic stock building for supermarkets and fruit quality control \cite{naranjo2020review}. In particular, we differ from the above amodal datasets in two main aspects: (1) We introduce for the first time a dataset containing extensive intra-class dense occlusion scenarios for amodal perception research. (2) Intra-AFruit is much larger than the above dataset and provides four types of annotations, including amodal instance masks, amodal visual appearance and dual-order relations. In contrast, the other datasets provide only some of these annotations (see Tab.~\ref{datasets}). As a large-scale dataset with diverse annotations, Intra-AFruit is promising for various amodal perception tasks.

\section{The Amodal Fruits Dataset}

Some amodal datasets \cite{zhu2017semantic, qi2019amodal} are created through human annotators. However, manual annotation is expensive, time-consuming and tends to provide inconsistent and inaccurate hallucinations for invisible regions. For example, different annotators may have different levels of detail and guesswork for invisible regions, and providing a complete appearance of a large number of occluded areas is extremely difficult. In contrast, synthetic images can provide a large number of accurate and detailed annotations at a low cost.

 \begin{figure}[t]
  \centering
   \includegraphics[width=\linewidth]{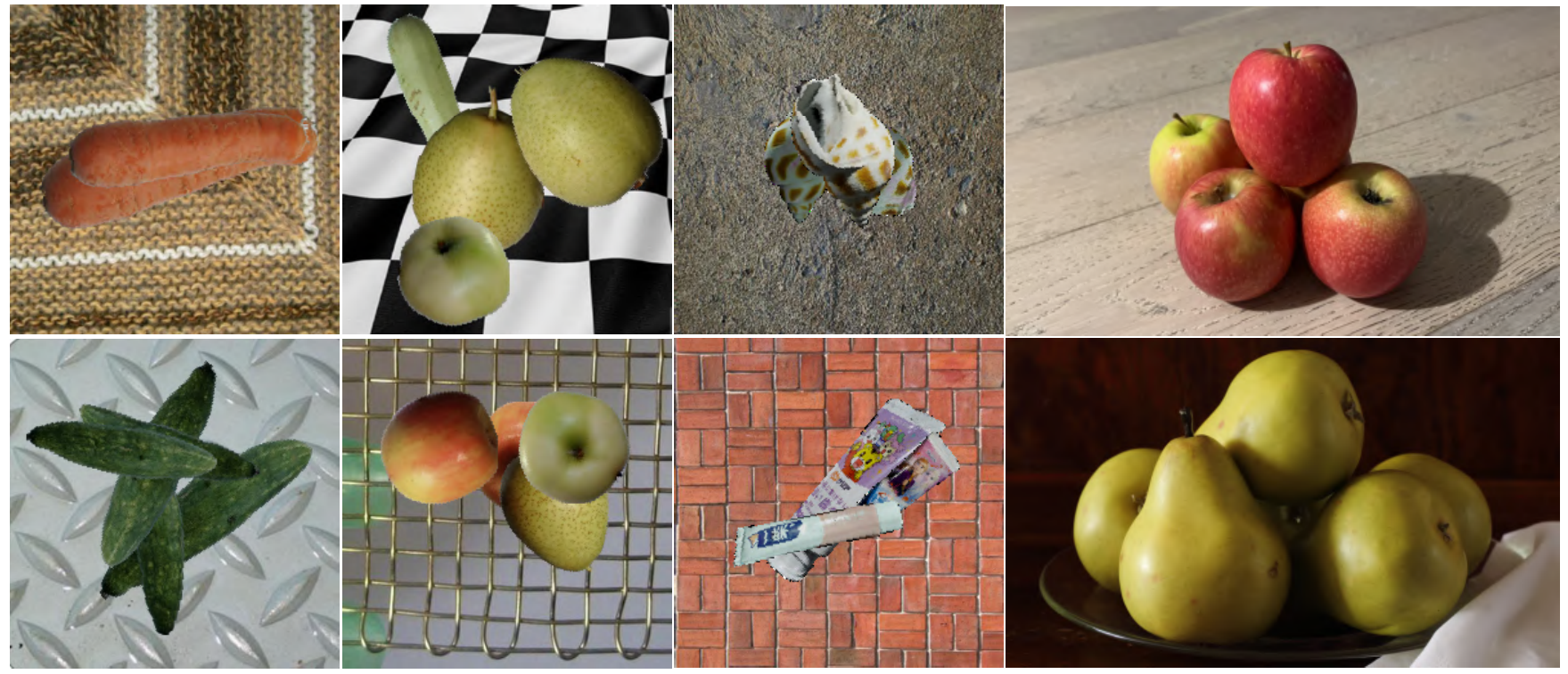}
   \caption{Example Images from our datasets: Intra-AFruit, Inter-AMix, ACom, and Real Images (ordered from left to right with two examples per column for each dataset). Intra-AFruit contains scenarios where instances of a single category occlude each other, while Inter-AMix considers the mutual occlusion of instances of different classes. On the other hand, ACom adds intra-class occlusion scenarios for other common object categories.}
   \label{examples images}
\end{figure}

We present a large-scale synthetic \textbf{Intra}-class \textbf{A}modal \textbf{Fruit}s dataset (Intra-AFruit). It is divided into two parts, with 210,000 images for training and 45,000 images for testing. Note that the instances in the training and test sets are from the training and test sets of the Fruits-360 \cite{murecsan2018fruit} respectively, so the instances in the test images are unseen in the training set. We also create a \textbf{A}modal \textbf{Com}mon objects (\textbf{ACom}) dataset to diversify the object categories beyond fruits and vegetables. We additionally created two datasets for testing purposes: an \textbf{Inter}-class \textbf{A}modal \textbf{Mix}ed Fruits dataset (Inter-AMix) and a real dataset (see Fig.~\ref{examples images}). In contrast to Intra-AFruit, Inter-AMix considers occlusion scenes of multiple categories of objects by randomly picking instances among all categories. As another part of the test, we collected a real-world image dataset of intra-class occlusion scenes containing 102 images and 317 instances. These real images are annotated by a human annotator based on estimating the occluded parts. In this section, we describe and analyse our datasets with informative statistics.

\begin{table*}[t]
    \footnotesize
    \centering
    \begin{tabular}{l|llll|llllll}
    \toprule
    & \multicolumn{4}{c|}{Fruits}                   & \multicolumn{6}{c}{Vegetables}                                     \\ 
Category & Golden  & Pear    & Pink Lady & Red Delicious & Zucchini & Carrot & Cabbage & Eggplant & Cucumber1 & Cucumber3 \\ \midrule
Num. in Train & 69195 &71522        &71479        &71701        &64498        &65065   & 69967     &57439        &69531        &64718        
\\ 
Num. in Test & 14847  & 15179 & 15220        & 15375  &   13904    &14022  &   15001   &12284&14832  &14077               
\\ \hline

Overall Ratio & 10.25\% & 10.58\% & 10.57\% &  10.62\% & 9.56\% & 9.65\% &10.36\%&  8.50\% & 10.29\% & 9.61\%\\ \bottomrule
    \end{tabular}
    \vspace*{-0.1cm}
    \caption{Categories distribution of Intra-AFruit. The number of instances in each category is nearly uniformly distributed in Intra-AFruit.}
    \label{Class}
\end{table*}

\subsection{Data Generation}
\textbf{Image Acquisition.}
 To generate the synthetic data, we use the processed images from the Fruits-360 \cite{murecsan2018fruit} as foreground objects and the images from the DTD \cite{cimpoi14describing} as backgrounds. Our dataset contains different fruits and vegetables with a total of 10 categories (see Tab.~\ref{Class}), and their labels follow the original Fruits-360‘s \cite{murecsan2018fruit} categories.

In creating Intra-AFruit, we considered the various challenges that can be faced in practical applications, such as highly occluded scenarios, objects of different sizes and angles, the location of different layers, and complex backgrounds. Concretely, each image contains between 2 and 5 instances with random rotation angles and sizes of a single category to ensure sufficient intra-class occlusion scenes in the dataset.
The background of each image is randomly sampled from the DTD \cite{cimpoi14describing} and resized to 256*256 pixels. The diversity of backgrounds facilitates the robustness of the trained model to different backgrounds. Image samples containing completely invisible instances (i.e., no visible pixels) are discarded, which follows the sense of human amodal perception \cite{chen2016amodal}, i.e., perceive the occluded region through the visible part of the object. After filtering, the number of preserved images reaches 255K, which we believe is sufficient for modelling such intra-class scenarios.

\textbf{Annotations.} 
As the Intra-AFruit instances are composited layer by layer onto the background, all annotations are automatically and accurately provided during the data generation process. The annotations consist of (1) Three types of instance masks, including visible, invisible and amodal masks. Multiple masks are essential for amodal methods that require various types of masks \cite{follmann2019learning, xiao2021amodal}; (2) Two types of order between instances. Occlusion order considers instances that share overlapping areas of amodal masks, including direct and indirect pairwise occlusion. Layer order, on the other hand, considers the relative distance to the camera. Instances in a cluster are labelled in order of proximity to distance, i.e., the layer of unoccluded objects is denoted as 0, then add 1 for instances directly occluded only by instances of layer 0, and so on. The different types of orders are considered to complement each other and facilitate the work of amodal order perception \cite{Lee_2022_CVPR}. For instance, an object may not occlude another object from a different layer; (3) The entire appearance of each instance and background, which provides the ground truth for the subsequent amodal appearance completion task. 

\begin{figure}[t]
  \centering
  \begin{subfigure}[t]{0.45\linewidth}
    \includegraphics[width=\linewidth]{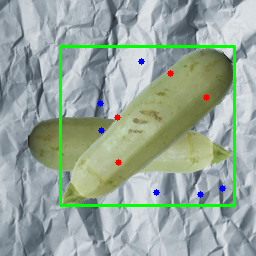}
    \caption{Annotation for Top Object}
    \label{fig:short-a}
  \end{subfigure}
  \hfill
  \begin{subfigure}[t]{0.45\linewidth}
    \includegraphics[width=\linewidth]{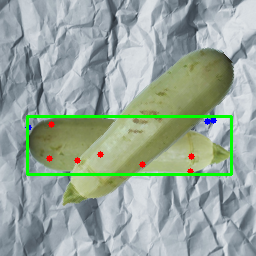}
    \caption{Annotation for Bottom Object}
    \label{fig:short-b}
  \end{subfigure}
  \vspace*{-0.2cm}
  \caption{An example of our point-based amodal instance annotation scheme for one of the instances in the image. For each instance, we randomly collected 10 points in its amodal bounding box (\textcolor{green}{green}) and labelled each point as an instance (\textcolor{red}{red}) if the point was within the amodal mask, while other sampled points in the box received the background labels (\textcolor{blue}{blue}). In this way points located within the entire object are treated as instances, even if part of the object is being occluded.}
  \label{point-based annotation}
\end{figure}

\begin{figure}[t]
\vspace*{-0.18cm}
  \centering
   \includegraphics[width=0.9\linewidth]{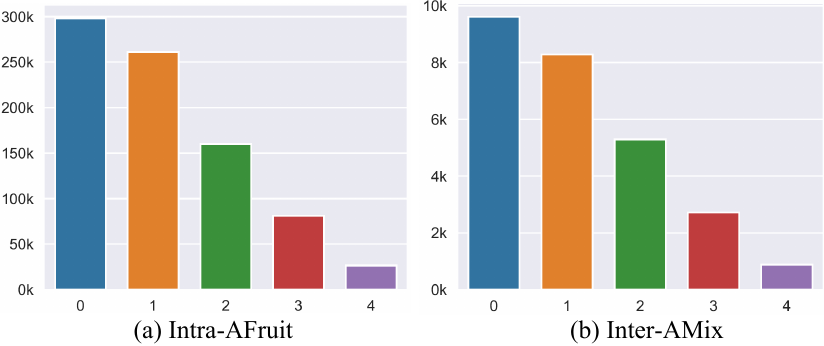}
   \vspace*{-0.1cm}
   \caption{Distribution of instances in each layer for (a) the Intra-AFruit dataset and (b) the Inter-AMix dataset. The numbers on the x-axis represent the layers. Depending on the occlusion of instances in the image, each image has up to five layers, where layer 0 represents fully visible instances and higher numbers indicate more deeply occluded layers.}
   \label{layer distribution}
\end{figure}

In addition, to support our point supervision approach, we generate point annotations based on the amodal mask. Specifically, for each instance, we use the instance‘s amodal bounding box and 10 randomly sampled points from the box, including the invisible parts. We annotate each point as either the object (even if it belongs to an invisible part) or the background (see Fig.~\ref{point-based annotation}).

We use the same pipeline to create ACom, except for using the processed images from OmniObject3D \cite{wu2023omniobject3d} as the foreground objects. ACom contains 10 categories of common objects, including anise, biscuit, bottle, candy, conch, donut, fire extinguisher, hammer, hat, and toothpaste.

\begin{figure*}[h]
  \centering
  \includegraphics[width=0.9\linewidth]{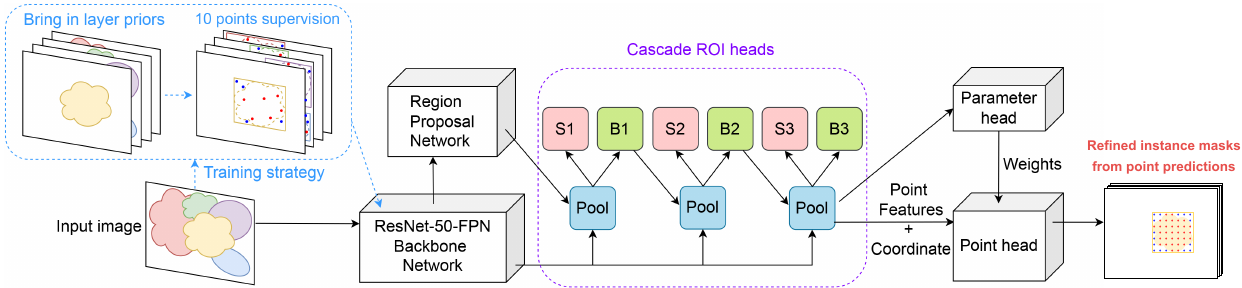}
   \caption{Overview of our network architecture. During training, we used only the 10-point annotation with the layer priors as supervision rather than the full mask. Cascade ROI heads include bounding box branches (denoted by "B") and mask branches (denoted by "S"). Note that box branches perform box regression and classification. The network generates a different weighted point head for each detected instance and then combines point features and coordinates for point predictions. Finally, the subdivision mask rendering algorithm \cite{kirillov2020pointrend} is used to refine the instance mask from point-wise predictions.}
   \label{architecture}
\end{figure*}

\subsection{Dataset Statistics}

Intra-AFruit consists of 255,000 images with 819,856 instances belonging to 10 categories, of which 675,115 instances are in the training set and 144,741 instances in the test set. Tab.~\ref{Class} shows the distribution of the instance categories. In addition to Intra-AFruit's training and test sets, we have kept a validation set containing 45,000 images and 144,542 instances for future use. The Inter-AMix dataset, which is used to test the model's generalisation performance, consists of 8,067 images with 26,508 instances. As shown in Fig.~\ref{layer distribution}, the instances of Intra-AFruit and Inter-AMix have similar distributions in the layers. ACom has 12,500 images and 45,485 instances in 10 categories, of which 36,276 instances are in the training set of 10,000 images, and 9,209 instances are in the test set of 2,500 images.

\section{Proposed Method}

Most modern instance segmentation methods with fully convolutional networks use non-maximal suppression (NMS) to remove densely overlapping proposal boxes for each class. This strategy, however, implicitly ignores scenes with highly occluded instances within classes and results in the inaccurate bounding box and mask predictions. Meanwhile, weakly supervised methods with the easier acquisition of ground truth annotations are urgently needed to make amodal segmentation applicable to a wide range of scenarios. To address these issues, we propose a conceptually simple but effective 
\textbf{P}ointly-supervised scheme with \textbf{L}ayer priors for amodal \textbf{In}tra-class instance segmentation (PLIn).

In this section, we describe the details of the proposed method, PLIn. Our goal is to predict the amodal mask for all instances of a given image. During training, the inputs to our model are the amodal bounding box, the 10 annotation points for each instance, and the layer order in which they are located. The simple strategy of bringing in layer priors allows the model to build separate representations for each layer and thus capture intra-class occlusions efficiently. Our approach follows a two-stage instance segmentation method of detecting before segmenting. Specifically, there are three major modules in PLIn: the backbone network, the Regional Proposal Network (RPN), and the Cascade ROI heads. Fig.~\ref{architecture} gives an overall view of PLIn. We follow the point supervision strategy of Implicit PointRend \cite{cheng2022pointly}, where different weighted point heads are generated for each detected instance, and then combined with point features and coordinates to make point predictions. As the last step, the subdivision mask rendering algorithm in \cite{kirillov2020pointrend} is used to refine instance masks with point predictions.

\subsection{Layer Priors}

Multiple overlapping intra-class instances in neighbouring ROIs may cause a failure of object recognition and instance segmentation. A primary reason for this is that many SOTA of segmentation methods, including Mask-RCNN \cite{he2017mask} and its variants \cite{cai2018cascade,follmann2019learning,kirillov2020pointrend,cheng2022pointly}, as well as some anchor-free methods \cite{lee2020centermask}, employ NMS in the inference process to remove dense overlapping proposal boxes. However, the classical NMS shows drawbacks in images containing dense occlusions because of the tendency to erroneously remove the boxes of highly occluded instances. This can be particularly fatal for amodal instance segmentation that focuses on densely occluded scenes and invisible regions. As an improvement, a popular NMS strategy performs NMS for each class independently so that overlapping boxes from different classes are less likely to be discarded mistakenly. This strategy, however, does not work well for scenarios containing a large number of highly occluded instances of the same class.

In order to mitigate this limitation, our proposed approach extends existing instance segmentation methods by adding layer priors, i.e. decoupling overlapping objects into multiple image layers. This allows the model to capture intra-class occlusion situations well during inference.

\begin{figure}[t]
  \centering
   \includegraphics[width=0.7\linewidth]{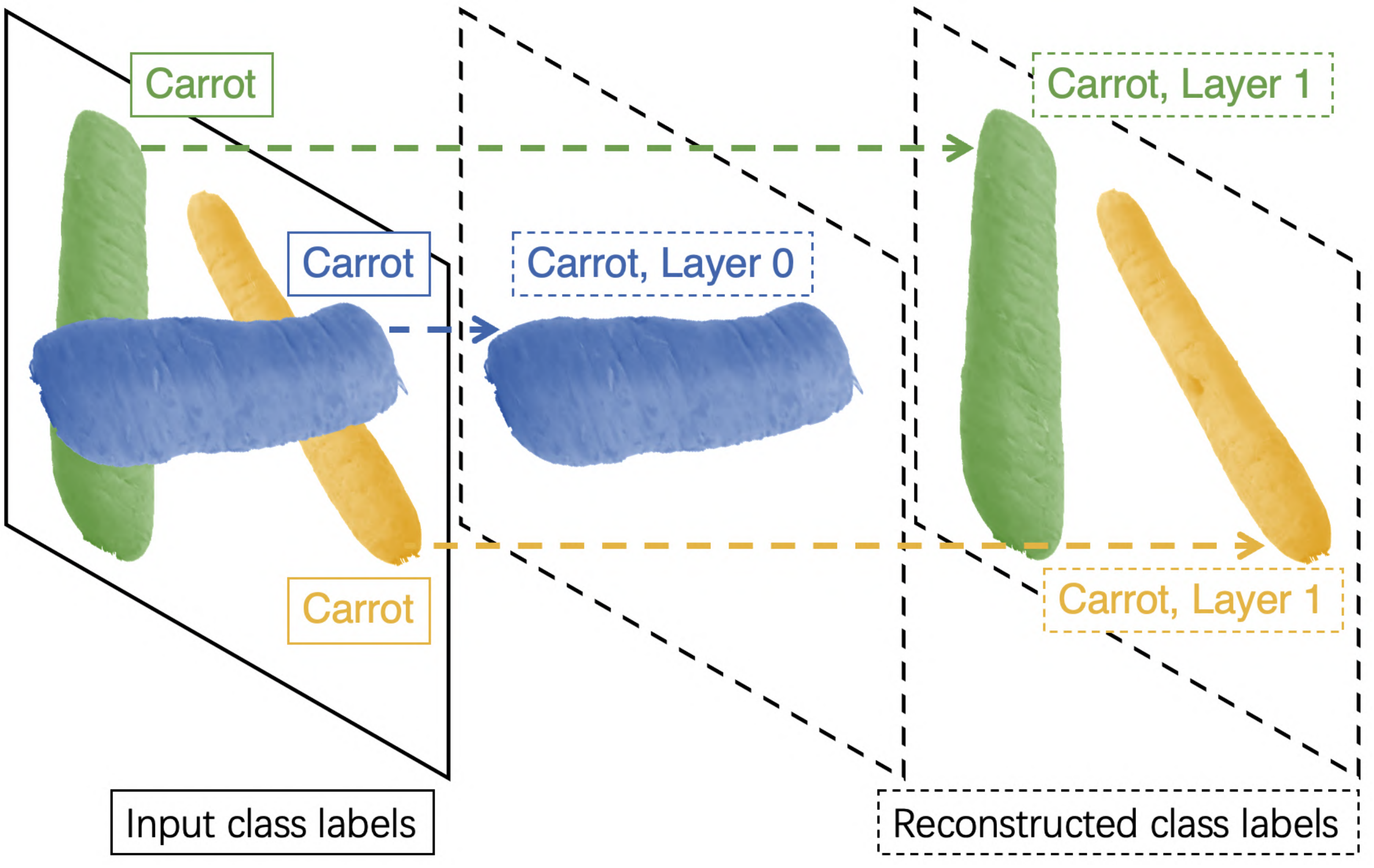}
   \vspace*{-0.1cm}
   \caption{Construction of the layer labels. Instances on layer 0 are fully visible, and instances on layer 1 are only occluded by instances on layer 0.}
   \label{layer priors}
\end{figure}

We use the layer priors to explicitly separate occlusion patterns by treating overlapping objects as instances of different layers. Let \( L(i) \) denote the layer of instance \( i \). The occlusion of instance \( i \) by any instance in set \( S_i \) (where \( S_i \) is the set of instances that occlude \( i \)) can be represented by the function \( O(i, j) \), where:
\begin{equation}
O(i, j) = 
\begin{cases} 
1, & \text{if } i \text{ is occluded by } j \\
0, & \text{otherwise.}
\end{cases}
\label{eq:occlusion}
\end{equation}
The layer order for instance \( i \) is defined as:
\begin{equation}
L(i) = 
\begin{cases} 
0, & \text{if } S_i = \emptyset \\
1 + \max_{j \in S_i} L(j), & \text{otherwise.}
\end{cases}
\label{eq:combined}
\end{equation}

Then, we construct $\{category, layer\}$ pairs for each instance as inputs to the model, and the network treats each $\{category, layer\}$ pair as an individual class, as shown in Fig.~\ref{layer priors}. Correspondingly, the model predicts a $\{category, layer\}$ pair for each detected instance. To produce a typical class-based prediction, we can add a simple post-processing step that ignores the predicted layer order. 

Since intra-class instances obscuring each other are defined as being in different layers, they are prevented from being erroneously removed during the NMS process. Furthermore, by introducing layer priors, the trained model has the ability to perceive the layer order for the instances simultaneously. 

\subsection{Point-based Supervision with Cascade Strategy}

Point-based annotations provide an alternative to the expensive per-pixel annotations typically required in instance segmentation algorithms by alleviating the need for supervision. Although efforts have been made to design point-based supervision for modal instance segmentation \cite{laradji2020proposal, cheng2022pointly}, there is still a research gap in extending to amodal instance segmentation.

We bring the point-supervised design for modal instance segmentation \cite{cheng2022pointly} to the amodal task and improve it for intra-class occlusion. By modifying the loss calculation of the mask branch, point-based supervision uses only the ground truth of points rather than the entire mask. Instead of training using matched regular grid labels from the full ground truth masks, point supervision uses bilinear interpolation to approximate the exact location of ground truth points. Once the predicted results and ground truth labels are available for the same points, the cross-entropy loss on the points can be applied, with the gradients propagated by bilinear interpolation.

However, the point-supervised approach may fail in intra-class occlusion scenarios due to multiple overlapping instances from the same category. As an improvement, we not only introduce layer priors to represent multilayer image structure, but also adopt the Cascade strategy which is more effective for intra-class overlap.

We utilised a 3-stage cascade training strategy \cite{cai2018cascade} that considers three IoU thresholds (0.5, 0.6, 0.7, respectively) instead of one to improve the quality of the detection. Detection is the first stage of our two-stage process, so a high-quality detector is crucial for subsequent segmentation. Cascade training uses the previous stage's output as input for the next stage, effectively adapting to the different input proposals. 

\subsection{Augmentation}
\label{sec:Augmentation}

\textbf{Point-based Augmentation.} The limited number of available data points may lead to a performance gap between point supervision and full supervision as training grows. To avoid this, we adopt the point-based data augmentation in \cite{cheng2022pointly}, where 5 points are randomly sub-sampled in each training iteration instead of using all 10 points.

\textbf{Test-Time Augmentation.} As opposed to most data augmentation techniques which augment data during the training phase, Test-Time Augmentation (TTA) is commonly used to generate more robust predictions in the testing phase \cite{krizhevsky2017imagenet, szegedy2015going, he2016deep}. Using TTA, the model gives a final output according to the average of the predictions made by multiple transformed versions of a given test image.

\section{Experiments}




\subsection{Experimental Setup}
\textbf{Datasets.} All experiments are conducted on our new datasets due to the lack of existing intra-class occlusion amodal datasets. We train models on the split of Intra-AFruit's training set (210K images) with the same settings and evaluate performance on the split of Intra-AFruit's test set. To further investigate the predictive capability of models in unseen scenes, we validate their generalization ability using real images and the Inter-AMix dataset containing inter-class occlusion scenes. By testing the models trained on our synthetic dataset on real images, we can assess their potential for real-world applications. Also, segmenting the amodal mask for Inter-AMix images is much more difficult as the model has only seen intra-class occlusion during training. 

\begin{figure}[t]
  \centering
   \includegraphics[width=\linewidth]{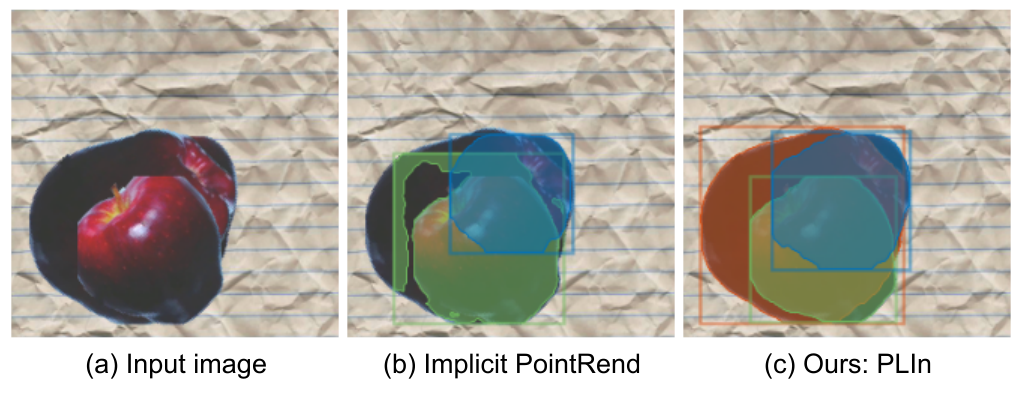}
    \vspace*{-0.8cm}
   \caption{Qualitative results comparison of the amodal mask predictions on (a) an Intra-AFruit test image by (b) Implicit PointRend \cite{cheng2022pointly} and (c) ours, both using the same experimental setup and point-based supervision, where PLIn detects all instances in the proximity ROIs and provides more reasonable amodal masks prediction. }
   \label{masks conflict}
\end{figure}

\begin{table}[t]
    \centering
    \footnotesize
    \begin{tabular}{lll|lll}
    \toprule
    \multicolumn{3}{c|}{Ablation Study}    & \multicolumn{3}{c}{\begin{tabular}[c]{@{}c@{}}Intra-class occlusion\\ Intra-AFruit test set\end{tabular}}  \\\midrule
LP & Cascade & TTA & $AP^{Box} \uparrow$ & $AP^{Mask} \uparrow$  & FPS $\uparrow$\\ \hline
 & & &71.984&74.995& 67.107
\\
$\checkmark$         &            &      &    80.656  &   84.202 & \textbf{67.204}    \\
       &     $\checkmark$    &         
&   82.892 &  80.911 & 57.768      \\
 $\checkmark$       &        $\checkmark$    &  & 88.668   &  86.187 & 22.126      \\
 $\checkmark$       & $\checkmark$           & $\checkmark$   &                        \textbf{88.722}    &       \textbf{87.410} & 2.841      \\                            
    \bottomrule
    \end{tabular}
    \vspace*{-0.1cm}
    \caption{Ablation study of our architectures. LP refers to Layer Priors. The top row is the baseline method, Implicit PointRend \cite{cheng2022pointly}. All models use a ResNet-50-FPN backbone \cite{he2016deep,lin2017feature}.} 
    \label{Ablation architectures}
\end{table}

\begin{table}[t]
    \centering
    \footnotesize
    \begin{tabular}{l|lllll}
    \toprule
    \multicolumn{1}{c|}{Ablation study} & \multicolumn{5}{c}{$AP^{Mask} \uparrow$ in classifying by layer order number} \\ 
& $L0$  & $L1$ & $L2$     & $L3$ & $L4$  \\ \midrule

 Layer Priors only &   92.710 & 78.325&52.682&29.079&16.212    \\

+ Cascade        &    94.432&80.831&56.487&  34.010& 19.551                                     \\
+ TTA   &   \textbf{95.487} & \textbf{83.176} & \textbf{58.208}& \textbf{35.330}& \textbf{20.821}      \\                            
    \bottomrule
    \end{tabular}
    \vspace*{-0.1cm}
    \caption{Performance by layer on the Intra-AFruit test set. We consider layer order rather than object category as the evaluation criterion. The fully visible layer 0 (L0) is predicted with the highest accuracy, while the deeper layers are less predictable. Nevertheless, the use of the Cascade training strategy brings better prediction performance to every layer.}
    \label{Ablation layer}
\end{table}

Note that the fully supervised methods are supervised with full amodal instance mask annotations, while the point supervised methods use only 10 labelled points.

\textbf{Evaluation Metrics.} We adopt the average precision (AP) of the standard COCO-style \cite{lin2014microsoft}, averaged over IoU thresholds from 0.5 to 0.95 with a step size of 0.05.

\subsection{Ablation Study} 
We follow the idea of collecting 10 points for each instance in \cite{cheng2022pointly}, which is a trade-off between annotator workload and model performance. As our model performs segmentation and layer order prediction simultaneously, in this section, we evaluate it from both perspectives.

We validate the effectiveness of different components proposed for PLIn. Tab.~\ref{Ablation architectures} presents quantitative comparisons of some ablation experiments performed on our Intra-AFruit dataset. To fairly compare the predictive power in instance amodal bounding boxes and masks, we ignore the predictive power for the instance layer here. We extend the existing SOTA point supervision approach for modal instance segmentation \cite{cheng2022pointly} to amodal intra-class instance segmentation by introducing layer priors and Cascade strategy. Both the introduction of the layer priors to modelling and the use of the Cascade strategy produced noticeable improvements compared to the baseline, where using only the layer priors obtained a better $AP^{Mask}$ and the only Cascade strategy achieved a better $AP^{Box}$. The results demonstrate that by incorporating the layer priors and the Cascade strategy, PLIn achieves better overall performance. In particular, as shown in Fig.~\ref{masks conflict}, for images with multiple overlapping objects of the same category in neighbouring ROIs, PLIn's design reduces mask conflict between highly overlapping instances. TTA also slightly improves performance.

Further exploration of the performance of the different layers based on layer priors is shown in Tab.~\ref{Ablation layer}. Instead of treating object categories as a basis for classification, we use the layer order of the instances without regard to item categories because we try to analyse the effect of the Cascade strategy and TTA on the performance of different layers. 

A typical failure mode we observed was unobstructed instances at layer 0 having the highest AP, while instances at deeper layers are harder to predict. This may be because the highly occluded objects are more challenging, but it's also worth noting that deeper layers have far fewer training instances (refer to Fig.~\ref{layer distribution}). From the results, we can conclude that the models using the Cascade strategy show better performance at every layer level, while the one using TTA as post-processing achieves the best performance at most layer levels. Overall, introducing the layer priors empowers the model to better predict the layer in which the instance falls. At the same time, further incorporating the Cascade training strategy and TTA post-processing improves the amodal segmentation results for all occlusion layer levels.

\begin{figure*}[t]
  \centering
   \includegraphics[width=\linewidth]{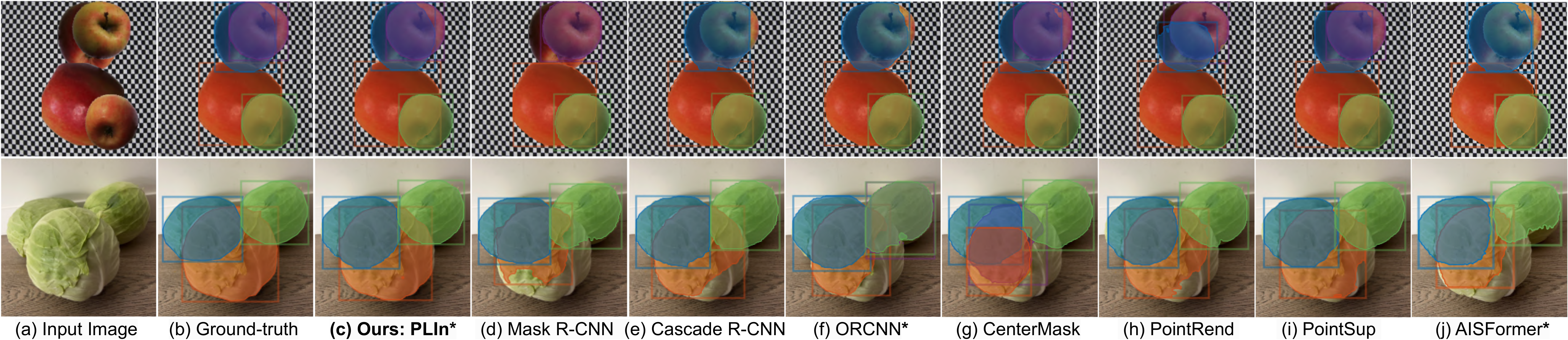}
   \captionsetup{skip=3pt}
   \caption{Qualitative results comparison of the amodal instance masks predictions on our Intra-AFruit test set (top) and a real-world image (bottom) by d) Mask R-CNN \cite{he2017mask}, e) Cascade R-CNN \cite{cai2018cascade},
   f) ORCNN \cite{follmann2019learning}, g) CenterMask \cite{lee2020centermask}, h) PointRend \cite{kirillov2020pointrend}, i) PointSup \cite{cheng2022pointly}, j) AISFormer \cite{Tran_2022_BMVC} and c) Ours. * Indicates methods designed specifically for amodal mask prediction.
   We show predictions with confidence greater than 0.5. We also give the estimated amodal bounding boxes as a reference for detection. Some methods fail to detect adjacent instances while others yield less reasonable estimates compared to ours. }
   \label{visual comparison}
\end{figure*}

\begin{table*}[t]
  \centering
  \scriptsize
  \begin{tabular}{l| c  | c c c c | c c | c c}
    \toprule
    Method & Supervision & \multicolumn{4}{c|}{Intra-AFruit test set} & \multicolumn{2}{c}{Inter-AMix} & \multicolumn{2}{c}{Real images} \\
    & (Init. designed*) &$AP^{Box} \uparrow$ & $AP^{Mask} \uparrow$ & $AP_{50}^{Mask} \uparrow$ & $AP_{75}^{Mask} \uparrow$ &  $AP^{Box} \uparrow$ & $AP^{Mask} \uparrow$  &  $AP^{Box} \uparrow$ & $AP^{Mask} \uparrow$ \\
    \midrule
    Mask R-CNN \cite{he2017mask}  &  full mask (Modal) &  72.068 & 70.022 & 90.428 & 76.991 & 62.904 & 61.420 & 42.645& 40.243\\
    Cascade R-CNN \cite{cai2018cascade}  &  full mask (Modal) & 81.694 & 74.598 & 89.871 & 82.311 & \textbf{71.629} & 65.221 & 54.464 & 48.010 \\
    ORCNN \cite{follmann2019learning}  & full mask (Amodal) & 73.381& 70.932 & 90.717 & 78.231 & 63.901 & 61.920 & 28.508 & 25.909\\ 
    CenterMask \cite{lee2020centermask} &full mask (Modal) & 78.296 & 76.315 & 95.211 & 82.039 & 61.982 &  60.313 & 39.583 & 37.422\\  
    PointRend \cite{kirillov2020pointrend}  &  full mask (Modal) & 72.461&76.194 &90.956&82.285& 63.409&66.399 & 45.494 & 46.445\\ 
    PointSup \cite{cheng2022pointly}  & 10 points (Modal) & 73.093 & 76.552 & 92.428 & 83.767 & 64.582 & 67.771 & 49.233 & 52.178 \\
    AISFormer \cite{Tran_2022_BMVC} & full mask (Amodal) & 73.745& 76.031 & 93.271& 84.520 & 62.530 & 63.394 & 33.858 & 33.705 \\ 
    \midrule
    PLIn (Ours) & 10 points (Amodal)  & \textbf{88.668}   &  \textbf{86.187}    &        \textbf{96.985}       &  \textbf{92.488} & 68.852 & \textbf{68.060} & \textbf{63.300} & \textbf{62.378}\\ 
    \bottomrule
  \end{tabular}
  \vspace*{-0.1cm}
  \caption{Comparison of our approach and other alternatives. PLIn using only 10-point supervision achieves the best results on the Intra-AFruit test set, and shows better generalisation than most above methods on the Inter-AMix test set. * Indicates whether a method was initially designed for amodal or modal instance segmentation.}
  \captionsetup{skip=2pt}
  \label{data comparison}
\end{table*}

\begin{table}[t]
    \scriptsize
    \centering
    \begin{tabular}{l|c|cc}
    \toprule
    Method & Supervision & \multicolumn{2}{c}{ACom test set} \\  
            &(Init. designed*) & \multicolumn{1}{c}{$AP^{Box} \uparrow$} & $AP^{Mask} \uparrow$ \\ \midrule
    BoxTeacher \cite{cheng2023boxteacher} & Weak (Modal) & \multicolumn{1}{c|}{55.942}      & 40.663      \\ 
    PointSup \cite{cheng2022pointly}& Weak (Modal) & \multicolumn{1}{c|}{48.680}      & 44.008      \\ 
    BoxInst \cite{tian2021boxinst} & Weak (Modal)& \multicolumn{1}{c|}{51.740}      & 23.408      \\ 
    AISFormer \cite{Tran_2022_BMVC} & Full (Amodal) & \multicolumn{1}{c|}{53.100} & 43.411 \\ 
    PointRend \cite{kirillov2020pointrend} & Full (Amodal) & \multicolumn{1}{c|}{49.019}      & 46.813   \\ 
    \midrule
    PLIn (Ours)   & Weak (Amodal) & \multicolumn{1}{c|}{\textbf{58.190}}      & \textbf{50.127} \\ \bottomrule
    \end{tabular}
    \vspace*{-0.1cm}
    \caption{Comparison of our approach and other alternatives on ACom. PLIn outperforms other methods on the ACom test set.}
    \label{newexperiments}
\end{table}

\subsection{Comparisons to Other Networks}

We compare our method with the SOTA instance segmentation methods with the same ResNet-50-FPN \cite{he2016deep,lin2017feature} backbone. All methods are trained on the Intra-AFruit training set with the same experimental setup and use ImageNet \cite{krizhevsky2017imagenet} pre-trained weights. TTA was omitted in all methods (including ours) as it is time-consuming but produced limited improvement (see Tab.~\ref{Ablation architectures}).

We present a qualitative comparison of existing instance segmentation methods on an Intra-AFruit test image with severe intra-class occlusion in Fig.~\ref{visual comparison}. In the top example, PLIn predicts more reasonable bounding boxes and shapes for the two instances at the top of the image, despite being heavily occluded from each other. In the bottom example, PLIn predicts more reasonable shapes for both fully visible and partially occluded objects in a realistic image.

As seen in Tab.~\ref{data comparison}, with the introduction of the layer priors, our weakly supervised method outperforms the best-available methods by more than 9.63\% in $AP^{Mask}$ and more than 6.97\% in $AP^{Box}$ on the Intra-AFruit test set of intra-class occlusion scenes. Regarding the generalisation ability tested on the unseen Inter-AMix dataset, the mask performance of PLIn is also optimal; although PLIn is second in $AP^{Box}$ to the fully supervised approach of \cite{cai2018cascade}, PLIn is able to simultaneously perform layer order perception while being trained using only 10-point supervision. PLIN also outperforms other methods by a large margin on the real images test set, proving that our weakly supervised approach has greater potential for generalisation to real images, where the annotation is expensive. We also conducted experiments on the ACom dataset to verify the performance of our approach on a broader range of objects beyond fruits and vegetables (see Tab.~\ref{newexperiments}).

\section{Conclusion}
We present new datasets and a weakly supervised strategy to solve the novel problem of amodal instance segmentation with heavy intra-class occlusion. The datasets with extensive annotation will facilitate future amodal completion tasks such as object order perception and inpainting. The proposed method effectively mitigates the negative impact of NMS on overlapping intra-class instances by introducing layer priors, while applying point supervision to amodal segmentation for the first time with an improved training strategy for scenarios of intra-class occlusion. Notably, experiments show that our method performs better than existing methods on our two synthetic intra-class occlusion datasets and real images. Meanwhile, we provide a benchmark for simultaneous amodal instance segmentation and layer order perception.

{\small
\bibliographystyle{ieee_fullname}
\bibliography{egbib}

\begin{thebibliography}{10}\itemsep=-1pt

\bibitem{back2022unseen}
Seunghyeok Back, Joosoon Lee, Taewon Kim, Sangjun Noh, Raeyoung Kang, Seongho Bak, and Kyoobin Lee.
\newblock Unseen object amodal instance segmentation via hierarchical occlusion modeling.
\newblock In {\em Proceedings of the International Conference on Robotics and Automation (ICRA)}. IEEE, 2022.

\bibitem{breitenstein2022amodal}
Jasmin Breitenstein and Tim Fingscheidt.
\newblock Amodal cityscapes: A new dataset, its generation, and an amodal semantic segmentation challenge baseline.
\newblock In {\em 2022 IEEE Intelligent Vehicles Symposium (IV)}, pages 1018--1025. IEEE, 2022.

\bibitem{cai2018cascade}
Zhaowei Cai and Nuno Vasconcelos.
\newblock Cascade r-cnn: Delving into high quality object detection.
\newblock In {\em Proceedings of the IEEE Conference on Computer Vision and Pattern Recognition (CVPR)}, pages 6154--6162, 2018.

\bibitem{chen2016amodal}
Siyi Chen, Hermann~J M{\"u}ller, and Markus Conci.
\newblock Amodal completion in visual working memory.
\newblock {\em Journal of Experimental Psychology: Human Perception and Performance}, 42(9):1344, 2016.

\bibitem{cheng2022pointly}
Bowen Cheng, Omkar Parkhi, and Alexander Kirillov.
\newblock Pointly-supervised instance segmentation.
\newblock In {\em Proceedings of the IEEE/CVF Conference on Computer Vision and Pattern Recognition (CVPR)}, pages 2617--2626, 2022.

\bibitem{cheng2023boxteacher}
Tianheng Cheng, Xinggang Wang, Shaoyu Chen, Qian Zhang, and Wenyu Liu.
\newblock Boxteacher: Exploring high-quality pseudo labels for weakly supervised instance segmentation.
\newblock In {\em Proceedings of the IEEE/CVF Conference on Computer Vision and Pattern Recognition (CVPR)}, pages 3145--3154, 2023.

\bibitem{cimpoi14describing}
M. Cimpoi, S. Maji, I. Kokkinos, S. Mohamed, and A. Vedaldi.
\newblock Describing textures in the wild.
\newblock In {\em Proceedings of the IEEE Conference on Computer Vision and Pattern Recognition (CVPR)}, 2014.

\bibitem{Cordts2016Cityscapes}
Marius Cordts, Mohamed Omran, Sebastian Ramos, Timo Rehfeld, Markus Enzweiler, Rodrigo Benenson, Uwe Franke, Stefan Roth, and Bernt Schiele.
\newblock The cityscapes dataset for semantic urban scene understanding.
\newblock In {\em Proc. of the IEEE Conference on Computer Vision and Pattern Recognition (CVPR)}, 2016.

\bibitem{ehsani2018segan}
Kiana Ehsani, Roozbeh Mottaghi, and Ali Farhadi.
\newblock Segan: Segmenting and generating the invisible.
\newblock In {\em Proceedings of the IEEE Conference on Computer Vision and Pattern Recognition (CVPR)}, pages 6144--6153, 2018.

\bibitem{follmann2019learning}
Patrick Follmann, Rebecca K{\"o}nig, Philipp H{\"a}rtinger, Michael Klostermann, and Tobias B{\"o}ttger.
\newblock Learning to see the invisible: End-to-end trainable amodal instance segmentation.
\newblock In {\em Proceedings of the IEEE Winter Conference on Applications of Computer Vision (WACV)}, pages 1328--1336. IEEE, 2019.

\bibitem{he2017mask}
Kaiming He, Georgia Gkioxari, Piotr Doll{\'a}r, and Ross Girshick.
\newblock Mask r-cnn.
\newblock In {\em Proceedings of the IEEE International Conference on Computer Vision (ICCV)}, pages 2961--2969, 2017.

\bibitem{he2016deep}
Kaiming He, Xiangyu Zhang, Shaoqing Ren, and Jian Sun.
\newblock Deep residual learning for image recognition.
\newblock In {\em Proceedings of the IEEE Conference on Computer Vision and Pattern Recognition (CVPR)}, pages 770--778, 2016.

\bibitem{hu2019sail}
Yuan-Ting Hu, Hong-Shuo Chen, Kexin Hui, Jia-Bin Huang, and Alexander~G Schwing.
\newblock Sail-vos: Semantic amodal instance level video object segmentation-a synthetic dataset and baselines.
\newblock In {\em Proceedings of the IEEE/CVF Conference on Computer Vision and Pattern Recognition (CVPR)}, pages 3105--3115, 2019.

\bibitem{kanizsa1979organization}
Gaetano Kanizsa.
\newblock {\em Organization in vision: Essays on Gestalt perception}.
\newblock Praeger Publishers, 1979.

\bibitem{ke2021deep}
Lei Ke, Yu-Wing Tai, and Chi-Keung Tang.
\newblock Deep occlusion-aware instance segmentation with overlapping bilayers.
\newblock In {\em Proceedings of the IEEE/CVF conference on computer vision and pattern recognition (CVPR)}, pages 4019--4028, 2021.

\bibitem{kirillov2020pointrend}
Alexander Kirillov, Yuxin Wu, Kaiming He, and Ross Girshick.
\newblock Pointrend: Image segmentation as rendering.
\newblock In {\em Proceedings of the IEEE/CVF Conference on Computer Vision and Pattern Recognition (CVPR)}, pages 9799--9808, 2020.

\bibitem{krizhevsky2017imagenet}
Alex Krizhevsky, Ilya Sutskever, and Geoffrey~E Hinton.
\newblock Imagenet classification with deep convolutional neural networks.
\newblock {\em Communications of the ACM}, 60(6):84--90, 2017.

\bibitem{laradji2020proposal}
Issam~H Laradji, Negar Rostamzadeh, Pedro~O Pinheiro, David Vazquez, and Mark Schmidt.
\newblock Proposal-based instance segmentation with point supervision.
\newblock In {\em Proceedings of the IEEE International Conference on Image Processing (ICIP)}, pages 2126--2130. IEEE, 2020.

\bibitem{Lee_2022_CVPR}
Hyunmin Lee and Jaesik Park.
\newblock Instance-wise occlusion and depth orders in natural scenes.
\newblock In {\em Proceedings of the IEEE/CVF Conference on Computer Vision and Pattern Recognition (CVPR)}, pages 21210--21221, June 2022.

\bibitem{lee2020centermask}
Youngwan Lee and Jongyoul Park.
\newblock Centermask: Real-time anchor-free instance segmentation.
\newblock In {\em Proceedings of the IEEE/CVF Conference on Computer Vision and Pattern Recognition (CVPR)}, pages 13906--13915, 2020.

\bibitem{lin2017feature}
Tsung-Yi Lin, Piotr Doll{\'a}r, Ross Girshick, Kaiming He, Bharath Hariharan, and Serge Belongie.
\newblock Feature pyramid networks for object detection.
\newblock In {\em Proceedings of the IEEE Conference on Computer Vision and Pattern Recognition (CVPR)}, pages 2117--2125, 2017.

\bibitem{lin2014microsoft}
Tsung-Yi Lin, Michael Maire, Serge Belongie, James Hays, Pietro Perona, Deva Ramanan, Piotr Doll{\'a}r, and C~Lawrence Zitnick.
\newblock Microsoft coco: Common objects in context.
\newblock In {\em Proceedings of the European Conference on Computer Vision (ECCV)}, pages 740--755. Springer, 2014.

\bibitem{mohan2022amodal}
Rohit Mohan and Abhinav Valada.
\newblock Amodal panoptic segmentation.
\newblock In {\em Proceedings of the IEEE/CVF Conference on Computer Vision and Pattern Recognition (CVPR)}, pages 21023--21032, 2022.

\bibitem{murecsan2018fruit}
Horea Mure{\c{s}}an and Mihai Oltean.
\newblock Fruit recognition from images using deep learning.
\newblock {\em Acta Universitatis Sapientiae, Informatica}, 10(1):26--42, 2018.

\bibitem{naranjo2020review}
Jos{\'e} Naranjo-Torres, Marco Mora, Ruber Hern{\'a}ndez-Garc{\'\i}a, Ricardo~J Barrientos, Claudio Fredes, and Andres Valenzuela.
\newblock A review of convolutional neural network applied to fruit image processing.
\newblock {\em Applied Sciences}, 10(10):3443, 2020.

\bibitem{nguyen2021weakly}
Khoi Nguyen and Sinisa Todorovic.
\newblock A weakly supervised amodal segmenter with boundary uncertainty estimation.
\newblock In {\em Proceedings of the IEEE/CVF International Conference on Computer Vision (CVPR)}, pages 7396--7405, 2021.

\bibitem{qi2019amodal}
Lu Qi, Li Jiang, Shu Liu, Xiaoyong Shen, and Jiaya Jia.
\newblock Amodal instance segmentation with kins dataset.
\newblock In {\em Proceedings of the IEEE/CVF Conference on Computer Vision and Pattern Recognition (CVPR)}, pages 3014--3023, 2019.

\bibitem{szegedy2015going}
Christian Szegedy, Wei Liu, Yangqing Jia, Pierre Sermanet, Scott Reed, Dragomir Anguelov, Dumitru Erhan, Vincent Vanhoucke, and Andrew Rabinovich.
\newblock Going deeper with convolutions.
\newblock In {\em Proceedings of the IEEE Conference on Computer Vision and Pattern Recognition (CVPR)}, pages 1--9, 2015.

\bibitem{tian2021boxinst}
Zhi Tian, Chunhua Shen, Xinlong Wang, and Hao Chen.
\newblock Boxinst: High-performance instance segmentation with box annotations.
\newblock In {\em Proceedings of the IEEE/CVF Conference on Computer Vision and Pattern Recognition (CVPR)}, pages 5443--5452, 2021.

\bibitem{Tran_2022_BMVC}
Minh~Q Tran, Khoa~HV Vo, Kashu Yamazaki, Arthur Fernandes, Michael~T Kidd, and Ngan Le.
\newblock Aisformer: Amodal instance segmentation with transformer.
\newblock In {\em 33rd British Machine Vision Conference 2022, {BMVC} 2022, London, UK, November 21-24}, 2022.

\bibitem{wu2023omniobject3d}
Tong Wu, Jiarui Zhang, Xiao Fu, Yuxin Wang, Jiawei Ren, Liang Pan, Wayne Wu, Lei Yang, Jiaqi Wang, Chen Qian, et~al.
\newblock Omniobject3d: Large-vocabulary 3d object dataset for realistic perception, reconstruction and generation.
\newblock In {\em Proceedings of the IEEE/CVF Conference on Computer Vision and Pattern Recognition (CVPR)}, pages 803--814, 2023.

\bibitem{xiao2021amodal}
Yuting Xiao, Yanyu Xu, Ziming Zhong, Weixin Luo, Jiawei Li, and Shenghua Gao.
\newblock Amodal segmentation based on visible region segmentation and shape prior.
\newblock In {\em Proceedings of the AAAI Conference on Artificial Intelligence}, volume~35, pages 2995--3003, 2021.

\bibitem{yan2019visualizing}
Xiaosheng Yan, Feigege Wang, Wenxi Liu, Yuanlong Yu, Shengfeng He, and Jia Pan.
\newblock Visualizing the invisible: Occluded vehicle segmentation and recovery.
\newblock In {\em Proceedings of the IEEE/CVF International Conference on Computer Vision (ICCV)}, pages 7618--7627, 2019.

\bibitem{zhan2020self}
Xiaohang Zhan, Xingang Pan, Bo Dai, Ziwei Liu, Dahua Lin, and Chen~Change Loy.
\newblock Self-supervised scene de-occlusion.
\newblock In {\em Proceedings of the IEEE/CVF Conference on Computer Vision and Pattern Recognition (CVPR)}, pages 3784--3792, 2020.

\bibitem{zhou2021human}
Qiang Zhou, Shiyin Wang, Yitong Wang, Zilong Huang, and Xinggang Wang.
\newblock Human de-occlusion: Invisible perception and recovery for humans.
\newblock In {\em Proceedings of the IEEE/CVF Conference on Computer Vision and Pattern Recognition (CVPR)}, pages 3691--3701, 2021.

\bibitem{zhu2017semantic}
Yan Zhu, Yuandong Tian, Dimitris Metaxas, and Piotr Doll{\'a}r.
\newblock Semantic amodal segmentation.
\newblock In {\em Proceedings of the IEEE Conference on Computer Vision and Pattern Recognition (CVPR)}, pages 1464--1472, 2017.

\end{thebibliography}
}

\end{document}